\documentclass{article}

 \usepackage[dblblindworkshop, final]{neurips_2025}
\workshoptitle{NPGML}

\usepackage[utf8]{inputenc} 
\usepackage[T1]{fontenc}    
\usepackage{hyperref}       
\usepackage{url}            
\usepackage{booktabs}       
\usepackage{amsfonts}       
\usepackage{nicefrac}       
\usepackage{microtype}      
\usepackage{xcolor}         
\usepackage{multirow}
\usepackage{graphicx}
\usepackage{caption}
\usepackage{subcaption}
\usepackage{algorithm}
\usepackage{algpseudocode}
\usepackage{booktabs}
\usepackage{wrapfig}
\usepackage{amsmath}
\usepackage{breakurl}
\usepackage{tikz}
\usetikzlibrary{positioning, arrows.meta, shapes.geometric, calc}

\title{Graph Representation Learning with Diffusion Generative Models}

\author{%
  Daniel Wesego \\
  Department of Computer Science\\
  University of Illinois Chicago\\
  Chicago, IL 60607 \\
  \texttt{dweseg2@uic.edu} \\
}

\usepackage{tcolorbox}

\begin{document}

\maketitle

\begin{abstract}
  Diffusion models have established themselves as state-of-the-art generative models across various data modalities, including images and videos, due to their ability to accurately approximate complex data distributions. Unlike traditional generative approaches such as VAEs and GANs, diffusion models employ a progressive denoising process that transforms noise into meaningful data over multiple iterative steps. This gradual approach enhances their expressiveness and generation quality. Not only that, diffusion models have also been shown to extract meaningful representations from data while learning to generate samples. Despite their success, the application of diffusion models to graph-structured data remains relatively unexplored, primarily due to the discrete nature of graphs, which necessitates discrete diffusion processes distinct from the continuous methods used in other domains. In this work, we leverage the representational capabilities of diffusion models to learn meaningful embeddings for graph data. By training a discrete diffusion model within an autoencoder framework, we enable both effective autoencoding and representation learning tailored to the unique characteristics of graph-structured data. We extract the representation from the combination of the encoder's output and the decoder's first time step hidden embedding. Our approach demonstrates the potential of discrete diffusion models to be used for graph representation learning. The code can be found at \url{https://github.com/DanielMitiku/Graph-Representation-Learning-with-Diffusion-Generative-Models}
\end{abstract}

\section{Introduction}

Representation learning is a central paradigm in modern machine learning, aiming to transform raw data into informative and compact representations that capture the underlying structure of the domain. Such representations enable a broad range of downstream tasks, including classification, clustering, and generation, by providing features that are more amenable to statistical modeling than the raw inputs ~\citep{representaion_bengio}. Advances in deep learning have driven rapid progress in different areas: variational autoencoders extract latent codes from images ~\citep{vae, rezende14}, transformer-based models learn contextualized embeddings of text ~\citep{attention_all_you_need}, and other areas such as robotics ~\citep{tereda2024efficient}. In the context of graphs, different forms of Graph Neural Networks (GNNs) produce representations of nodes, edges, and entire graphs that preserve structural and feature information ~\citep{graph_representation}. These methods have established representation learning as a unifying principle across domains such as computer vision, natural language processing, multimodal data, and graph analysis ~\citep{representation_learning1, wesego2024revising, representation_learning2}.

Graphs are powerful structures for modeling relationships between entities, and they are widely used in domains such as social networks, biological networks, transportation systems, and knowledge graphs~\citep{graph_survey}. In these domains, graphs represent complex systems, where nodes typically correspond to entities, and edges represent relationships or interactions between them. The ability to analyze and extract insights from graph-structured data is critical for applications such as recommendation systems, drug discovery, fraud detection, social network analysis ~\citep{graph_app1, gat}, and others. Graph representation learning has therefore become essential, aiming to map graph-structured data into low-dimensional vector embeddings that preserve both structure and features. These embeddings facilitate efficient analysis and downstream tasks such as node classification, link prediction, and graph classification~\cite{gcn}. Despite significant progress, challenges remain in graph representation learning. Graphs often exhibit highly complex structures, with varying node degrees, long-range dependencies, and hierarchical relationships that are difficult to compress into low-dimensional embeddings~\citep{gcn}. Furthermore, real-world graphs are frequently heterogeneous, containing multiple types of nodes, edges, and attributes that must be modeled within a unified framework~\citep{hetrogenous_graphs}. Dynamic graphs introduce an additional challenge, where evolving structures require methods that can adapt representations to capture temporal changes~\citep{dynamic_graphs}.

Parallel to these developments, recent advances in deep generative models, particularly diffusion models, have opened new directions for representation learning \citep{preechakul2021diffusion}. Diffusion models have achieved remarkable success in high-quality sample generation across domains such as image, audio, and video synthesis \citep{diff_ho_etal, any-to-any-codi, wesego}, and are beginning to be explored in the context of graphs ~\citep{digress}. Their iterative denoising process naturally learns hierarchical and expressive latent structures, suggesting that diffusion models can provide powerful graph representations. However, the use of diffusion models for graph representation learning remains at an early stage, with several open challenges.

In this work, we investigate discrete diffusion autoencoders for graph representation learning. Discrete diffusion models are particularly well-suited for graph data, where node and edge features are often categorical or discrete in nature. By leveraging their generative capabilities, we aim to enhance the quality and expressiveness of graph embeddings, while enabling unsupervised learning in settings where labeled data is scarce or costly. We evaluate our framework on benchmark datasets, including the Protein and IMDB-B dataset from TUDatasets~\cite{TUDatasets}, and demonstrate its effectiveness in downstream tasks. The application of diffusion models to graph learning holds significant promise ~\cite{yang2023directionaldiffusionmodelsgraph}. Beyond improved embeddings, their generative nature allows the synthesis of novel graphs, which is especially valuable in applications such as molecular graph generation and drug discovery~\cite{drug_discovery}. The central hypothesis is that the strong generative capacity of diffusion models requires learning compressed, informative representations that can serve as robust graph embeddings, as proven in other modalities.

To achieve this, we make use of diffusion autoencoders, which extend a standard diffusion model into an autoencoder framework, where an encoder network learns representations and a diffusion decoder reconstructs the input data conditioned on the output of the encoder \citep{preechakul2021diffusion}. Discrete diffusion autoencoders, in particular, are tailored for discrete data, making them especially suitable for graphs~\cite{d3pm}, and in this paper, we explore their potential for graph representation learning. Figure~\ref {fig:diffusion_gae} shows the overall framework that is used to get the embeddings after training. Our contributions are summarized as follows:

\begin{itemize}
\item \textbf{Framework.} We introduce a discrete diffusion graph autoencoder (DDGAE) for graph-structured data. Our model leverages discrete diffusion processes to progressively denoise graph inputs, enabling the capture of complex structural patterns and dependencies.
\item \textbf{Representation learning.} We show how discrete diffusion models can improve the quality of graph representations by transforming the discrete nature of graph structures into a latent embedding. By integrating diffusion models with an autoencoder architecture, DDGAE learns compact, expressive graph embeddings. The final representation combines the encoder output with the embeddings of the diffusion decoder, enhancing representational richness. Unlike standard diffusion models that rely on repetitive sampling, our approach requires only a single-step sampling from the diffusion decoder during inference, while naturally supporting unsupervised learning and reducing reliance on labeled data.
\item \textbf{Empirical validation.} We conduct extensive experiments on different graph benchmark datasets from TUDatasets, demonstrating that DDGAE achieves superior performance in downstream graph classification tasks compared to strong baselines.
\end{itemize}

\begin{figure}[ht]
    \centering
    \resizebox{\linewidth}{!}{%
    \begin{tikzpicture}[
        node distance=0.8cm and 1.0cm,
        every node/.style={align=center},
        input/.style={rectangle, draw=yellow!70, fill=yellow!20, rounded corners, minimum width=2.0cm, minimum height=0.8cm},
        encoder/.style={rectangle, draw=blue!70, fill=blue!10, rounded corners, minimum width=2.0cm, minimum height=0.8cm},
        decoder/.style={rectangle, draw=red!70, fill=red!5, rounded corners,
                        minimum width=3.4cm, minimum height=1.6cm, thick},
        embedding/.style={ellipse, draw=green!70, fill=green!10, minimum width=2.2cm, minimum height=0.9cm, thick},
        tasks/.style={rectangle, draw=purple!70, fill=purple!10, rounded corners, minimum width=2.2cm, minimum height=0.9cm},
        arrow/.style={-{Stealth[scale=1]}, thick},
    ]

    \node (graph) [input] {Input Graph \\ (Adjacency + Features)};
    \node (encoder) [encoder, right=of graph] {GCN Encoder \\ $\mathbf{h}_{enc}$};
    \node (decoder) [decoder, below=1.0cm of encoder] {};
    \node (embedding) [embedding, right=1.0cm of decoder] {Embedding \\ $\mathbf{z} = [\mathbf{h}_{enc}, \mathbf{h}_{dec_{int}}]$};
    \node (tasks) [tasks, right=1.0cm of embedding] {Downstream \\ Tasks};

    \node at ($(decoder.north)+(0,-0.25)$) {\scriptsize Diffusion Decoder};

    \node[rectangle, draw=red!70, fill=red!40, minimum width=0.4cm, minimum height=0.4cm] (d1) at ($(decoder.center)+(-1.0,0)$) {};
    \node[rectangle, draw=red!70, fill=red!25, minimum width=0.4cm, minimum height=0.4cm] (d2) at ($(decoder.center)+(-0.3,0)$) {};
    \node[rectangle, draw=red!70, fill=red!15, minimum width=0.4cm, minimum height=0.4cm] (d3) at ($(decoder.center)+(0.4,0)$) {};
    \node[rectangle, draw=red!70, fill=red!5,  minimum width=0.4cm, minimum height=0.4cm] (d4) at ($(decoder.center)+(1.1,0)$) {};

    \draw[arrow, red!70] (d1) -- (d2);
    \draw[arrow, red!70] (d2) -- (d3);
    \draw[arrow, red!70] (d3) -- (d4);

    \draw[arrow] (graph) -- (encoder);
    \draw[arrow] (graph.south) .. controls +(down:0.4cm) and +(-0.5,0cm) .. (decoder.west);

    \draw[thick] (encoder.south) -- ++(0,-0.5) coordinate (split); 
    \draw[arrow] (split) -- (decoder.north);
    \draw[arrow] (split) -| (embedding.north west);

    \draw[arrow] (decoder.east) -- (embedding.west);
    \draw[arrow] (embedding.east) -- (tasks.west);

    \end{tikzpicture}%
    }
    \caption{Discrete Diffusion Graph AutoEncoder (DDGAE) embedding extraction: The trained encoder extracts features, which are concatenated with the intermediate output of the trained diffusion decoder as the final embedding $\mathbf{z}$ that will be used for different downstream tasks.}
    \label{fig:diffusion_gae}
\end{figure}

\section{Related Works}

\subsection{Diffusion Models}

Diffusion models have emerged as powerful generative models, achieving strong performance across domains including image synthesis, molecule generation, and representation learning \citep{diff_ho_etal, digress, yang2023directionaldiffusionmodelsgraph, wesego2024revising}. These models operate by progressively adding noise to data in the forward diffusion process and learning to reverse this process. This iterative framework enables the generation of high-quality samples that closely resemble the target data distribution. Denoising Diffusion Probabilistic Models (DDPMs) introduced the foundational framework for diffusion-based generation, demonstrating their ability to produce realistic images through iterative denoising~\citep{diff_ho_etal}. Improved DDPMs further enhanced this framework by optimizing noise schedules, learning the variance, and refining architectural designs~\citep{dhariwal}. Other notable advancements include Stable Diffusion, which integrates text conditioning with diffusion models in the latent space to generate high-resolution, text-guided images~\citep{rombach2021highresolution}.

\subsection{Discrete Diffusion Models for Graphs}

Discrete diffusion models extend diffusion principles to discrete data, making them suitable for domains such as text, categorical attributes, and graphs. Unlike continuous diffusion models, which operate in a continuous state space, discrete diffusion models handle data in a way that respects the inherent discreteness of the input \citep{d3pm, argmax_hoogeboom}. When applied to graphs, these models leverage diffusion processes to capture the complex relationships and hierarchical structures inherent in graph-structured data. Since graphs are inherently discrete, most diffusion models for graphs operate directly in the discrete space \citep{digress, edge}. DiGress, a discrete denoising diffusion model, generates graphs with categorical node and edge attributes, demonstrating strong effectiveness in handling graph-structured data~\citep{digress}. \citet{lggm} further expanded DiGress by training it on graphs from multiple domains to improve generalization. EDGE~\citep{edge} is another discrete diffusion model that generates adjacency matrices conditioned on degree distributions. By using an absorbing distribution of empty graphs as the terminal state, EDGE reduces the number of diffusion steps, effectively addressing graph sparsity. Together, these approaches highlight the growing potential of discrete diffusion models for structured data generation.

\subsection{Graph Representation Learning}

Graph representation learning has been a central research area, focusing on transforming graph-structured data into low-dimensional embeddings that preserve structural and semantic information. Several approaches have been proposed to tackle the challenges of learning meaningful graph representations, including contrastive, generative, and autoencoder-based methods. GraphCL (Graph Contrastive Learning) introduced a self-supervised framework that leverages graph augmentations to maximize agreement between representations of the same graph under different transformations~\citep{graphcl}. GraphMAE (Graph Masked Autoencoder) adapts masked autoencoding, widely used in NLP, to graphs by masking portions of the input and training the model to reconstruct them, thereby learning structural patterns~\citep{graphmae}. GraphVAE (Graph Variational Autoencoder) is a generative framework that models the probabilistic distribution of graph data; it learns latent variables representing graph structures and attributes to compress and reconstruct graphs~\citep{graphvae}. More recently, \citet{yang2023directionaldiffusionmodelsgraph} applied diffusion models to graph representation learning by extracting outputs from intermediate layers and injecting directional noise.

In this study, we propose a \textbf{Discrete Diffusion Graph Autoencoder (DDGAE)} for graph representation learning. Specifically, we use discrete diffusion models as decoders over adjacency matrices and a GCN encoder to extract latent representations $\mathbf{z}$. Training is performed solely through the discrete diffusion process in the decoder, with gradients propagated back to the encoder. To enrich the learned representations, we combine the encoder output with intermediate embeddings from the diffusion decoder to form the final representation $\mathbf{z}$.

\section{Methodology}

This section describes the methodology of our \textbf{Discrete Diffusion Graph Autoencoder (DDGAE)} for graph-structured data. We first review the general discrete diffusion framework and then detail its application within our graph autoencoder.

\subsection{Discrete Diffusion Framework}

Discrete diffusion models define a Markov chain of $T$ diffusion steps that progressively corrupt the input data into noise. Let $\mathbf{x}_0$ denote the original data and $\mathbf{x}_T$ denote the fully noised state. The forward process is governed by transition probabilities $q(\mathbf{x}_t | \mathbf{x}_{t-1})$, designed so that the distribution of $\mathbf{x}_t$ approaches a tractable noise distribution as $t$ increases. A generative model $p_\theta(\mathbf{x}_{t-1} | \mathbf{x}_t)$ is then trained to reverse this process, iteratively denoising from $\mathbf{x}_T$ to reconstruct the original data $\mathbf{x}_0$.

The marginal probability of the data is defined as $p_\theta(\mathbf{x}_0) := \int p_\theta(\mathbf{x}_{0:T}) d\mathbf{x}_{1:T}$, where the joint distribution over the diffusion trajectory is  
$p_\theta(\mathbf{x}_{0:T}) := p_\theta(\mathbf{x}_T) \prod_{t=1}^T p_\theta(\mathbf{x}_{t-1} | \mathbf{x}_t)$, and the forward process is $q(\mathbf{x}_{1:T} | \mathbf{x}_0) := \prod_{t=1}^T q(\mathbf{x}_t | \mathbf{x}_{t-1})$. Training maximizes the evidence lower bound (ELBO) of the log-likelihood \citep{d3pm}:
\begin{align}
    \mathbb{E}[-\log p_\theta(\mathbf{x}_0)] \leq \mathbb{E}_q \Big[-\log \frac{p_\theta(\mathbf{x}_{0:T})}{q(\mathbf{x}_{1:T} | \mathbf{x}_0)}\Big].
\end{align}

This can be expressed as a sum of KL divergences over diffusion steps:
\begin{equation}
\begin{aligned}
    L_{VB} = 
    & \mathbb{E}_{q(\mathbf{x}_0)} \Big[ \underbrace{D_{KL}[q(\mathbf{x}_T | \mathbf{x}_0) || p_\theta(\mathbf{x}_T)]}_{L_T} \Big] + \\
    & \sum_{t=2}^T \mathbb{E}_{q(\mathbf{x}_t | \mathbf{x}_0)} \Big[ \underbrace{D_{KL}[q(\mathbf{x}_{t-1} | \mathbf{x}_t, \mathbf{x}_0) || p_\theta(\mathbf{x}_{t-1} | \mathbf{x}_t)]}_{L_{t-1}} \Big] - \mathbb{E}_{q(\mathbf{x}_1 | \mathbf{x}_0)} \Big[ \underbrace{\log p_\theta(\mathbf{x}_0 | \mathbf{x}_1)}_{L_0} \Big].
\end{aligned}
\end{equation}

Here, $L_T$ contains no trainable parameters and is zero by design. The final training objective is a linear combination of the ELBO and a term directly predicting $\mathbf{x}_0$, controlled by a hyperparameter $\lambda$ ~\citep{d3pm}:
\begin{equation}
\label{eq:main_diff_loss}
    L_\lambda = L_{VB} + \lambda \mathbb{E}_{q(\mathbf{x}_0)} \mathbb{E}_{q(\mathbf{x}_t | \mathbf{x}_0)} \left[ -\log \tilde{p}_\theta(\mathbf{x}_0 | \mathbf{x}_t) \right].
\end{equation}

\subsection{Discrete Diffusion Graph Autoencoder}

We apply the discrete diffusion framework to graph-structured data. The input graph is denoted $\mathbf{G}(\mathbf{X}, \mathbf{A})$, where $\mathbf{X}$ contains node features and $\mathbf{A}$ is the adjacency matrix. An encoder network $E_\phi(\mathbf{G})$ maps the input graph to a latent representation $\mathbf{z}_{enc}$. While any graph neural network can be used, we adopt a GCN to capture both structural and feature information effectively.  

The decoder is a discrete diffusion model operating on the adjacency matrix $\mathbf{A}$, conditioned on the encoder representation $\mathbf{z}_{enc}$. During training, the adjacency matrix undergoes a forward diffusion process, producing a noisy version $\mathbf{A}_t$ according to a predefined schedule that transforms $\mathbf{A}$ into an absorbing state of zeros. The decoder, $p_\theta(\mathbf{A}_{t-1} | \mathbf{A}_t, \mathbf{z}_{enc})$, iteratively reconstructs $\mathbf{A}$ from the noisy state, step-by-step, conditioned on $\mathbf{z}_{enc}$. The reconstructed adjacency matrix is denoted $\mathbf{\hat{A}}$.  

The final graph representation $\mathbf{z}$ is obtained by concatenating the encoder embedding $\mathbf{z}_{enc}$ with an intermediate embedding from the UNet-based diffusion decoder. After training, a single pass through the model suffices. First, we obtain $\mathbf{z}_{enc}$ forwarding the data through the encoder; this is passed to the decoder along with $\mathbf{A}_0$ to obtain the intermediate embedding, which is concatenated with $\mathbf{z}_{enc}$ to produce the final embedding $\mathbf{z}$ used for downstream tasks.

\section{Experiments}

This section outlines the experimental setup of our proposed model and the baselines. We compare our model against relevant baselines on a graph classification task using the PROTEINS and IMDB-BINARY datasets.

\subsection{Dataset}

The PROTEINS dataset comprises 1113 graphs representing proteins, each classified as either an enzyme (class 1) or non-enzyme (class 0). The graphs have an average of 39 nodes, with each node representing an amino acid and edges representing interactions between them. The IMDB-BINARY dataset consists of 1000 ego-networks extracted from the Internet Movie Database (IMDB), where each graph represents the collaboration network of actors in a movie. The graphs are classified into two categories based on the genre of the movie. Nodes correspond to actors, and edges indicate co-appearances in the same movie. Each graph has an average of 19 nodes and 193 edges. These datasets provide a suitable benchmark for evaluating our model's ability to learn meaningful representations from graph-structured data \citep{TUDatasets}.

\subsection{Baselines}

We compare our discrete diffusion graph autoencoder (DDGAE) against multiple baseline models used in ~\citep{yang2023directionaldiffusionmodelsgraph}, including 
Infograph ~\citep{sun2019infograph}, GraphCL ~\citep{graphcl}, JOAO ~\citep{you2021graph}, GCC ~\citep{qiu2020gcc},  MVGRL ~\citep{hassani2020contrastive},
2020), GraphMAE ~\citep{graphmae}, and DDM ~\citep{yang2023directionaldiffusionmodelsgraph}. From supervised learning methods, the comparisons include GIN ~\citep{xu2018powerful}.

\subsection{Model Architecture and Training}

Our model utilizes a Graph Convolutional Network (GCN) encoder to extract the latent representation $\mathbf{z}_{enc}$ from the input graph's features and adjacency matrix. The node embeddings from the GCN are aggregated using mean pooling to obtain a graph-level embedding. The decoder employs a UNET architecture, commonly used in diffusion models ~\citep{preechakul2021diffusion}, to reconstruct the adjacency matrix from the latent representation and noise. A diffusion timestep of 32 was used for the discrete diffusion process, and a latent size of 64 + 64 = 128 dimensions is used across our model. The models are trained on a single Nvidia GPU T4. The training takes approximately 1 hour for each dataset.

\subsection{Evaluation}

To evaluate the quality of the learned graph representations, we adopt the procedure by ~\citet{yang2023directionaldiffusionmodelsgraph} where we first extract the representations from the models and train an SVM classifier using 10-fold cross-validation on the extracted representations $\mathbf{z}$. We use classification accuracy as an evaluation metric on how good the learned representations are across the different models. This evaluation scheme allows us to directly assess the effectiveness of the learned representations in capturing discriminative information for graph classification from the graph properties.

\subsection{Results}

Table~\ref{tab:graph_results} presents the main results reporting the accuracy of the models trained on the representations learned by each model. The baselines used are similar to ~\citet{yang2023directionaldiffusionmodelsgraph}, and we evaluated our model similarly to have a fair comparison.

\begin{table}[ht]
\centering
\caption{Results of supervised (top 2) and unsupervised representation learning for graph classification datasets}
\label{tab:graph_results}
\begin{tabular}{lcc}
\toprule
Dataset & IMDB-B & PROTEINS \\
\midrule
GIN         & 75.1$\pm$5.1 & 76.2$\pm$2.8 \\
\midrule
Infograph   & 73.03$\pm$0.87 & 74.44$\pm$0.31  \\
GraphCL     & 71.14$\pm$0.44 &  74.39$\pm$0.45 \\
JOAO        & 70.21$\pm$3.08 &  74.55$\pm$0.41\\
GCC         & 72             & -   \\
MVGRL       & 74.20$\pm$0.70 & -   \\
GraphMAE    & 75.52$\pm$0.66 & 75.30$\pm$0.39  \\

DDM         & {76.40$\pm$0.22} & 75.47$\pm$0.50 \\
\midrule
\textbf{DDGAE} & \textbf{76.90$\pm$0.03} & \textbf{76.28$\pm$0.05} \\
\bottomrule
\end{tabular}
\end{table}

Our DDGAE model achieves the highest test accuracy on both datasets, demonstrating the superiority of the learned representations compared to the baseline models. This result highlights the effectiveness of our approach in capturing the complex structural information within graph data, leading to more discriminative and informative representations, opening a new research path towards using discrete diffusion autoencoder models for graph representation learning.

\section{Conclusion and Discussion}

In this paper, we introduce a discrete diffusion graph autoencoder model (DDGAE) for learning representations of graph-structured data. Our approach leverages the power of discrete diffusion models to capture the complex dependencies within the graph nodes and edges. By combining this generative framework with an encoder network, we learn a latent representation that effectively captures the underlying structure of the graph data. This representation can be used for various downstream tasks, such as graph generation, classification, and other use cases. 

Despite the promising results, our approach has some limitations. First, the computational cost of training diffusion-based decoders on large graphs can be significant. Second, our current model primarily focuses on static graphs with categorical node and edge features, and does not directly handle dynamic graphs. For future work, we plan to extend DDGAE to address these limitations by exploring more efficient diffusion schedules, incorporating heterogeneous and temporal graph data, and evaluating the model across a wider variety of datasets and tasks. Additionally, we aim to investigate the integration of contrastive or self-supervised objectives to further enhance the quality of learned representations. Overall, we believe that discrete diffusion autoencoders offer a promising new direction for graph representation learning, and our work lays the foundation for further exploration in this area.

\bibliography{bibl}
\bibliographystyle{plainnat}


\newpage

\section*{NeurIPS Paper Checklist}

\begin{enumerate}

\item {\bf Claims}
    \item[] Question: Do the main claims made in the abstract and introduction accurately reflect the paper's contributions and scope?
    \item[] Answer: \answerYes{} 
    \item[] Justification: Our claims match the results of our experiments

\item {\bf Limitations}
    \item[] Question: Does the paper discuss the limitations of the work performed by the authors?
    \item[] Answer: \answerYes{} 
    \item[] Justification: Limitation discussed in the conclusion section

\item {\bf Theory assumptions and proofs}
    \item[] Question: For each theoretical result, does the paper provide the full set of assumptions and a complete (and correct) proof?
    \item[] Answer: \answerNA{} 
    \item[] Justification: Our equations are based on previous papers, prrofs can be found on those papers.

    \item {\bf Experimental result reproducibility}
    \item[] Question: Does the paper fully disclose all the information needed to reproduce the main experimental results of the paper to the extent that it affects the main claims and/or conclusions of the paper (regardless of whether the code and data are provided or not)?
    \item[] Answer: \answerYes{} 
    \item[] Justification: Information is available to reproduce the results, plus see attached code

\item {\bf Open access to data and code}
    \item[] Question: Does the paper provide open access to the data and code, with sufficient instructions to faithfully reproduce the main experimental results, as described in supplemental material?
    \item[] Answer: \answerYes{} 
    \item[] Justification: Attached code has everything to reproduce the results.

\item {\bf Experimental setting/details}
    \item[] Question: Does the paper specify all the training and test details (e.g., data splits, hyperparameters, how they were chosen, type of optimizer, etc.) necessary to understand the results?
    \item[] Answer: \answerYes{} 
    \item[] Justification: Yes, described in the experiment section; other details can be found in the attached code.

\item {\bf Experiment statistical significance}
    \item[] Question: Does the paper report error bars suitably and correctly defined or other appropriate information about the statistical significance of the experiments?
    \item[] Answer: \answerYes{} 
    \item[] Justification: Experiment has standard deviation for the different k-fold values. Couldn't retrain the models again due to computational resources, but have attached the code.

\item {\bf Experiments compute resources}
    \item[] Question: For each experiment, does the paper provide sufficient information on the computer resources (type of compute workers, memory, time of execution) needed to reproduce the experiments?
    \item[] Answer: \answerYes{} 
    \item[] Justification: Yes, please look at the experiment section
    
\item {\bf Code of ethics}
    \item[] Question: Does the research conducted in the paper conform, in every respect, with the NeurIPS Code of Ethics \url{https://neurips.cc/public/EthicsGuidelines}?
    \item[] Answer: \answerYes{} 
    \item[] Justification: We respect and follow the Neurips Code of Ethics

\item {\bf Broader impacts}
    \item[] Question: Does the paper discuss both potential positive societal impacts and negative societal impacts of the work performed?
    \item[] Answer: \answerNA{} 
    \item[] Justification: Not required for our work
    
\item {\bf Safeguards}
    \item[] Question: Does the paper describe safeguards that have been put in place for responsible release of data or models that have a high risk for misuse (e.g., pretrained language models, image generators, or scraped datasets)?
    \item[] Answer: \answerNA{} 
    \item[] Justification: Not required for our work

\item {\bf Licenses for existing assets}
    \item[] Question: Are the creators or original owners of assets (e.g., code, data, models), used in the paper, properly credited and are the license and terms of use explicitly mentioned and properly respected?
    \item[] Answer: \answerYes{} 
    \item[] Justification: Datasets used are cited correctly.

\item {\bf New assets}
    \item[] Question: Are new assets introduced in the paper well documented and is the documentation provided alongside the assets?
    \item[] Answer: \answerNA{} 
    \item[] Justification: No new assets introduced

\item {\bf Crowdsourcing and research with human subjects}
    \item[] Question: For crowdsourcing experiments and research with human subjects, does the paper include the full text of instructions given to participants and screenshots, if applicable, as well as details about compensation (if any)? 
    \item[] Answer: \answerNA{} 
    \item[] Justification: Not used in our work

\item {\bf Institutional review board (IRB) approvals or equivalent for research with human subjects}
    \item[] Question: Does the paper describe potential risks incurred by study participants, whether such risks were disclosed to the subjects, and whether Institutional Review Board (IRB) approvals (or an equivalent approval/review based on the requirements of your country or institution) were obtained?
    \item[] Answer: \answerNA{} 
    \item[] Justification: Not required for our work

\item {\bf Declaration of LLM usage}
    \item[] Question: Does the paper describe the usage of LLMs if it is an important, original, or non-standard component of the core methods in this research? Note that if the LLM is used only for writing, editing, or formatting purposes and does not impact the core methodology, scientific rigorousness, or originality of the research, declaration is not required.
    \item[] Answer: \answerNA{} 
    \item[] Justification: Doesn't involve LLMs

\end{enumerate}

\end{document}